\documentclass[sigconf]{acmart}
\usepackage{enumitem}
\usepackage{multirow}
\usepackage{algorithm}
\usepackage{algorithmic}
\usepackage{balance}
\usepackage{colortbl}
\AtBeginDocument{%
  }

\copyrightyear{2026}
\acmYear{2026}
\setcopyright{cc}
\setcctype{by}
\acmConference[MM '26]{Proceedings of the 34th ACM International Conference on Multimedia}{November 10--14, 2026}{Rio de Janeiro, Brazil}
\acmBooktitle{Proceedings of the 34th ACM International Conference on Multimedia (MM '26), November 10--14, 2026, Rio de Janeiro, Brazil}
\acmDOI{10.1145/3767308.3835463}
\acmISBN{979-8-4007-2213-4/2026/11}
\begin{document}
\title{VISOR: Agentic Visual Retrieval-Augmented Generation via Iterative Search and Over-horizon Reasoning}


\author{Yucheng Shen}
\authornote{These authors contributed equally to this work.}
\affiliation{%
  \institution{Soochow University}
  \city{Suzhou}
  \state{Jiangsu}
  \country{China}}
\email{ycshensudaer@stu.suda.edu.cn}

\author{Jiulong Wu}
\authornotemark[1]
\affiliation{%
  \institution{Baidu Inc.}
  \city{Beijing}
  \country{China}}
\email{wjlwujiulong@gmail.com}

\author{Jizhou Huang}
\affiliation{%
  \institution{Baidu Inc.}
  \city{Beijing}
  \country{China}}
\email{hjizhou@qq.com}

\author{Dawei Yin}
\affiliation{%
  \institution{Baidu Inc.}
  \city{Beijing}
  \country{China}}
\email{yindawei@acm.org}

\author{Lingyong Yan}
\authornote{Corresponding authors.}
\affiliation{%
  \institution{Baidu Inc.}
  \city{Beijing}
  \country{China}}
\email{lingyongy@gmail.com}

\author{Min Cao}
\authornotemark[2]
\affiliation{%
  \institution{Soochow University}
  \city{Suzhou}
  \state{Jiangsu}
  \country{China}}
\email{mcao@suda.edu.cn}

\renewcommand{\shortauthors}{Yucheng Shen et al.}

\begin{abstract}

Visual Retrieval-Augmented Generation (VRAG) empowers Vision-Language Models to retrieve and reason over visually rich documents. To tackle complex queries requiring multi-step reasoning, agentic VRAG systems interleave reasoning with iterative retrieval. However, existing agentic VRAG faces two critical bottlenecks.
(1) Visual Evidence Sparsity: key evidence is scattered across pages yet processed in isolation, hindering cross-page reasoning; moreover, fine-grained intra-image evidence often requires precise visual actions, whose misuse degrades retrieval quality;
(2) Search Drift in Long Horizons: the accumulation of visual tokens across retrieved pages dilutes context and causes cognitive overload, leading agents to deviate from their search objective.
To address these challenges, we propose VISOR (Visual Retrieval-Augmented Generation via Iterative Search and Over-horizon Reasoning), a unified single-agent framework. VISOR features a structured Evidence Space for progressive cross-page reasoning, coupled with a Visual Action Evaluation and Correction mechanism to manage visual actions. Additionally, we introduce a Dynamic Trajectory with Sliding Window and Intent Injection to mitigate search drift. They anchor the evidence space while discarding earlier raw interactions, preventing context from being overwhelmed by visual tokens.
We train VISOR using a Group Relative Policy Optimization-based Reinforcement Learning (GRPO-based RL) pipeline with state masking and credit assignment tailored for dynamic context reconstruction. Extensive experiments on ViDoSeek, SlideVQA, and MMLongBench demonstrate that VISOR achieves state-of-the-art performance while maintaining reasonable and controllable computational costs for long-horizon visual reasoning tasks. The source code is available at \url{https://github.com/syc1336/VISOR}.
\end{abstract}




\maketitle

\section{Introduction}

Although Language Models (LMs) and Vision-Language Models (VLMs) have advanced rapidly in recent years and demonstrated remarkable capabilities across a wide range of tasks~\cite{achiam2023gpt, bai2023qwen, liu2023visual, liu2024deepseek, bai2025qwen3}, they remain constrained by fixed training data and are therefore prone to hallucination and knowledge gaps. Retrieval-Augmented Generation (RAG)~\cite{lewis2020retrieval,liu2025visual} mitigates these issues by grounding model outputs in externally retrieved evidence. Notably, real-world documents often contain rich visual content, such as charts, tables, and figures, which cannot be faithfully preserved through conventional text extraction. To bridge this gap, Visual RAG (VRAG) extends RAG to the visual domain~\cite{faysse2024colpali,yu2024visrag,jiang2024mmsearch}, enabling models to retrieve and reason directly over images rendered from document pages.

However, these VRAG works typically follow a fixed retrieve-then-read pipeline that performs only \textbf{a single round} of retrieval, leaving them unable to gather the additional evidence needed when a question requires reasoning across multiple steps~\cite{wang2025vrag}. For this, \textbf{agentic} VRAG is proposed~\cite{yao2022react,jin2025search,wu2025mmsearch,wang2025vrag}, and interleaves reasoning with iterative retrieval actions, so as to dynamically collect evidence across multiple turns. Notably, recent work~\cite{wang2025vrag} enriches the agent’s action space with visual actions, such as crop-and-zoom, enabling it to selectively magnify specific regions of a retrieved image (e.g., a chart segment or table cell) for fine-grained analysis.

\begin{figure}[htbp]
\centering
\includegraphics[width=0.92\columnwidth]{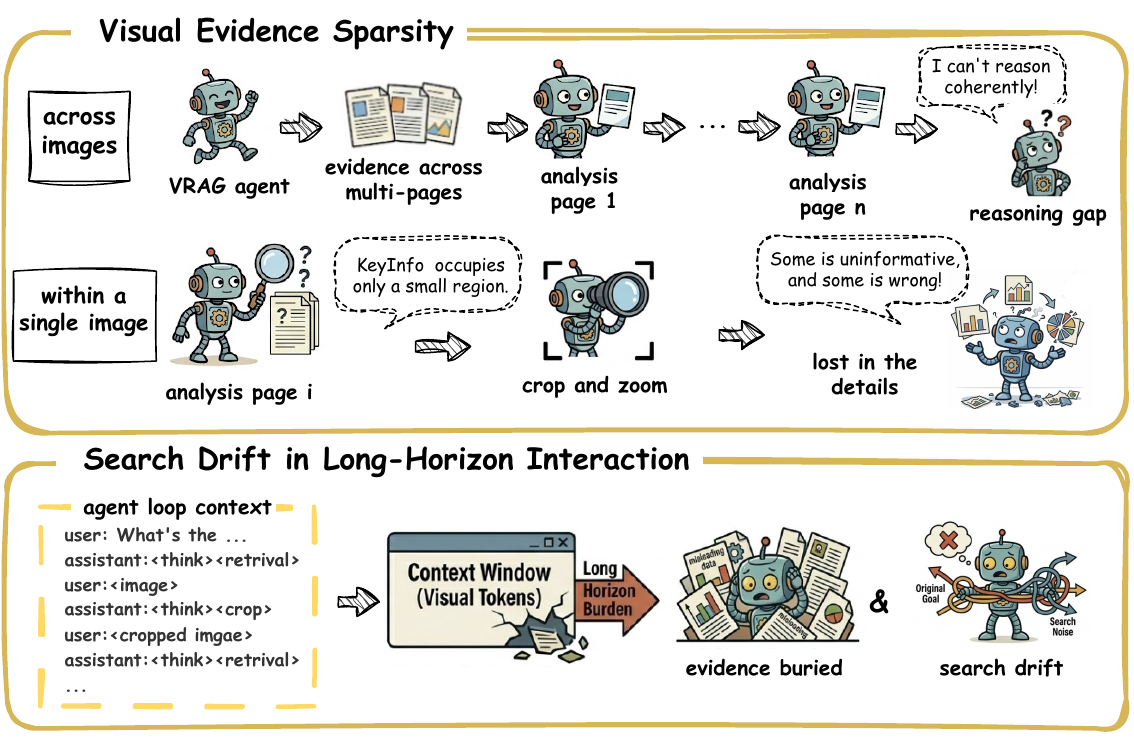}
\caption{Two critical bottlenecks in agentic VRAG:
(Top)\textbf{Visual Evidence Sparsity}, with relevant clues scattered across pages or small image regions;
(Bottom)\textbf{Long-Horizon Search Drift}, where growing visual context obscures prior evidence and the original goal.}
\label{fig:main}
\end{figure}

Despite this progress, they still face two critical bottlenecks when handling complex questions over visually rich content in long-horizon scenarios, as illustrated in Figure~\ref{fig:main}.
(1) \textbf{Visual Evidence Sparsity.}
Answering complex queries often requires locating and integrating key evidence from multiple document images, which is hindered by sparsity across two granularities. 
\emph{Across images}, relevant clues are typically scattered across multiple pages; however, existing systems process each retrieved page in isolation and lack mechanisms for cross-page evidence accumulation, leaving partial observations fragmented and insufficient to support coherent multi-step reasoning. 
\emph{Within a single image}, query-relevant information often occupies only a small region, e.g., specific cells in a dense table or a particular data series in a chart. Extracting such fine-grained evidence necessitates precise visual actions like crop-and-zoom. Yet, without an effective evaluation mechanism to guide these actions, agents risk performing suboptimal operations: cropping irrelevant regions introduces distracting content, or redundant zooms consume precious interaction turns without yielding useful signals.
(2) \textbf{Search Drift in Long-Horizon Interaction.}
This challenge is significantly amplified in the visual domain. Each retrieved document image consumes thousands of visual tokens, far exceeding the cost of textual content, causing the context window to saturate rapidly during multi-turn interactions. This leads to two compounding issues: first, previously gathered evidence becomes buried under accumulating visual data, making it difficult for the model to locate, or reuse earlier findings, ultimately undermining the coherence and completeness of the final answer; second, as visual tokens increasingly dominate the context, the agent’s attention drifts away from the original query, resulting in retrievals and actions that progressively deviate from user intent.

To address these bottlenecks, we propose \textbf{VISOR}, a \textbf{V}isual Retrieval Augmented Generation via
\textbf{I}terative \textbf{S}earch and \textbf{O}ver-horizon \textbf{R}easoning, illustrated in Figure~\ref{fig:overview}.
First, to mitigate \textbf{visual evidence sparsity}, VISOR introduces a \emph{structured Evidence Space} that explicitly accumulates query-relevant observations across retrieval iterations. This enables joint cross-page reasoning, moving beyond isolated per-page processing. Also, a \emph{Visual Action Evaluation and Correction} mechanism is developed to assess the utility and outcome of each visual action, thereby pruning noisy or irrelevant crops and selectively incorporating validated evidence into the Evidence Space.
Second, to overcome \textbf{search drift in long-horizon interaction}, VISOR incorporates two key mechanisms: \emph{Dynamic Trajectory}, which maintains a real-time, up-to-date Evidence Space pinned at the top of the context window, while recent interactions are retained via a sliding window; and \emph{Intent Injection}, which re-anchors the agent to the original user query at every retrieval step to preserve goal fidelity. This design effectively bounds context growth without sacrificing critical information.
Notably, VISOR adopts an agent-based loop architecture and is thus trained end-to-end: first via Supervised Fine-Tuning (SFT), followed by GRPO-based Reinforcement Learning (RL)~\cite{shao2024deepseekmath}. The reward function is carefully crafted to jointly optimize retrieval precision and final answer correctness.
Extensive experiments on established benchmarks, ViDoSeek~\cite{wang2025vidorag}, SlideVQA~\cite{tanaka2023slidevqa}, and MMLongBench~\cite{ma2024mmlongbench}, show that VISOR consistently outperforms existing agentic baselines even without task-specific fine-tuning, and its performance further improves with SFT+RL training.

Our main contributions are summarized as follows:
\begin{itemize}[leftmargin=*, nosep]
\item We propose \textbf{VISOR}, an agentic framework for long-horizon Visual RAG that addresses visual evidence sparsity and search drift via structured evidence accumulation, visual action correction, dynamic trajectory, and intent injection.

\item We design a two-stage training pipeline tailored for VISOR's dynamic context, with a carefully curated SFT dataset and a reward function that jointly optimizes retrieval precision and answer correctness.

\item Extensive experiments on ViDoSeek, SlideVQA, and MMLongBench demonstrate that VISOR consistently outperforms strong fine-tuned baselines on both Qwen2.5-VL-7B and 3B.
\end{itemize}

\section{Related Work}
\subsection{Visual Retrieval-Augmented Generation}
RAG has demonstrated significant advantages in addressing knowledge intensive problems~\cite{lewis2020retrieval,gao2023retrieval,yang2024crag}, with traditional text-based methods retrieving relevant passages for answer generation. However, with the widespread adoption of visually rich documents such as slides, reports, and scanned PDFs, knowledge is no longer confined to plain text. Early approaches rely on OCR or document parsing to extract textual content from images~\cite{huang2022layoutlmv3}, but such pipelines are lossy and fail to preserve layout, chart, and figure information. Recently, OCR-free retrieval methods~\cite{kim2022ocr} have emerged that directly align textual queries with document page images: ColPali~\cite{faysse2024colpali} introduced late-interaction visual retrieval via token-level similarity between query text and image patch embeddings; VisRAG~\cite{yu2024visrag} further explored generation pipelines operating directly on retrieved visual content. Recent work also explores multimodal RAG systems~\cite{jiang2024mmsearch} combining visual and textual information for more accurate retrieval and reasoning. EVisRAG~\cite{sun2025visrag} introduced evidence-based reasoning over multiple retrieved images to support multi-image understanding. Our work builds upon these visual retrieval foundations and further strengthens multi-image understanding.

\subsection{Agentic RAG with Reinforcement Learning}
The agentic paradigm for RAG was pioneered by ReAct~\cite{yao2022react}, which interleaves reasoning with actions so that models can decide when and what to retrieve. Reinforcement learning has emerged as an effective approach for improving reasoning in language models~\cite{guo2025deepseek,jaech2024openai,shao2024deepseekmath}, and recent work has extended it to VLMs for visual reasoning tasks~\cite{liu2025visual,wu2025facial}. In the agentic RAG setting, Search-R1~\cite{jin2025search} first applied RL to train language models for text-based agentic retrieval, and its visual extension Search-R1-VL adapted the framework to multimodal scenarios; R1-Router~\cite{peng2025learning} trained a routing policy to dynamically decide retrieval actions; MMSearch-R1~\cite{wu2025mmsearch} and WebWatcher~\cite{geng2025webwatcher} extended RL-based agents to multi-modal search and richer tool sets; VRAG-RL~\cite{wang2025vrag} defined a visual perception action space with crop-and-zoom operations and introduced multi-turn RL training for VRAG. However, these works do not address the challenges of multi-image evidence management and semantic drift in long-horizon visual reasoning. One natural direction to tackle these issues decomposes the task across specialized agents: ViDoRAG~\cite{wang2025vidorag} proposed actor-critic iterative reasoning with separate agents for planning, retrieval, and answering; SLEUTH~\cite{liu2025resolving} and M3RAG~\cite{du2026m3rag} decomposed the task into specialized modules. However, this decomposition can not support end-to-end optimization, as each component is trained or prompted independently with misaligned objectives, and sacrifices efficiency by routing all queries through the full pipeline regardless of complexity. Our work instead addresses these challenges within a single unified agent loop, achieving the benefits of both RL-based end-to-end training and robust long-horizon multi-image reasoning.

\section{Method}

\begin{figure*}[t]
\centering
\includegraphics[width=0.72\textwidth]{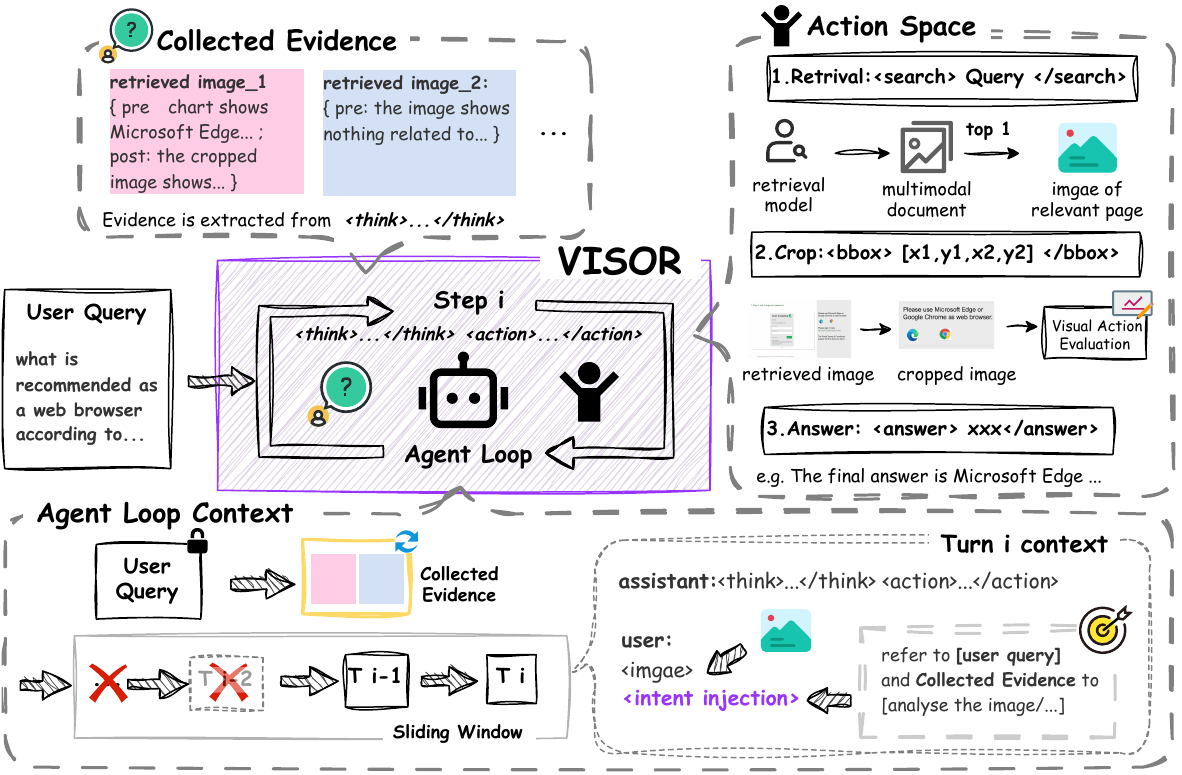}
\caption{Overview of \textbf{VISOR}. At each step $i$, the agent produces a $\langle$\texttt{think}$\rangle$\ldots$\langle$\texttt{action}$\rangle$ response. Evidence extracted from the reasoning trace is accumulated in a structured \emph{Evidence Collection} space $\mathcal{E}$. The action space comprises three operations: \texttt{search}, \texttt{crop}, and \texttt{answer}. The agent loop context is reconstructed at each turn: the user query and $\mathcal{E}$ are always pinned at the top, while only the last $W$ turns of raw interactions are retained via a \emph{Sliding Window}; each turn's observation also includes an \emph{Intent Injection} prompt to re-anchor the agent's focus.
}
\label{fig:overview}
\end{figure*}

\subsection{Task Formulation}

Given a natural language query $q$ and a large corpus $\mathcal{C} = \{I_1, I_2, \ldots, I_N\}$ comprising $N$ page-level images sourced from visually rich documents (\emph{e.g.}, slides, reports), the goal is to generate a final answer $a$ by iteratively retrieving and reasoning over relevant images. In practice, $\mathcal{C}$ is heterogeneous: pages from documents of varying topics and formats are pooled into a single flat index devoid of document-level boundaries. Consequently, answering a query often necessitates synthesizing evidence scattered across multiple pages within this mixed corpus.

\subsection{VISOR Agent Framework}\label{sec:overview}

As illustrated in Figure~\ref{fig:overview} and Algorithm~\ref{alg:visor}, \textbf{VISOR} is an iterative agentic framework wherein a single agent interleaves retrieval, visual reasoning, and evidence accumulation to perform visual RAG.
At each turn, the agent first reflects on its accumulated observations within a $\langle$\texttt{think}$\rangle$ block. It then issues one of three actions—\texttt{search}, \texttt{crop}, or \texttt{answer}—and subsequently receives a new observation from the environment. The loop continues until the agent either generates a final answer or reaches the maximum allowable number of turns. The framework relies on three core design components that respectively govern how the agent reasons, acts, and manages its context across turns.

\begin{algorithm}[t]
\caption{VISOR Agent Loop}\label{alg:visor}
\begin{algorithmic}[1]
\REQUIRE Query $q$, corpus $\mathcal{C}$
\REQUIRE Max turns $T$, sliding window size $W$
\STATE Initialize evidence space $\mathcal{E} \leftarrow \emptyset$, history $\mathcal{H} \leftarrow [\,]$
\STATE \textit{// $o_t$: environment observation returned after each action (retrieved image $I_k$ or cropped region $I'$, bundled with the intent-injected prompt)}
\STATE Construct initial prompt $\mathbf{P}$ from $q$
\FOR{$t = 1$ \TO $T$}
    \STATE \textit{// Dynamic Trajectory Sliding Window (\S\ref{sec:overview})}
    \STATE Construct context $\mathbf{C}_t \leftarrow [\,\mathbf{P};\; \mathcal{E};\; \text{last } W \text{ turns of } \mathcal{H}\,]$
    \STATE Generate response $\mathbf{r}_t$: reasoning $\theta_t$ (in $\langle$\texttt{think}$\rangle$ tags) followed by action $a_t(c_t)$, $a_t \in \{$\texttt{search}, \texttt{crop}, \texttt{answer}$\}$ \textit{($c_t$ is the action content: a search query, crop coordinates, or final answer string, respectively)}
    \STATE \textit{// Evidence Space update}
    \STATE Extract evidence from $\theta_t$; update $\mathcal{E}[I_k]$ for current page $I_k$
    \IF{$a_t = $ \texttt{answer}}
        \RETURN $c_t$ as the final answer
    \ELSIF{$a_t = $ \texttt{search}}
        \STATE Retrieve images from $\mathcal{C}$ using query $c_t$
        \STATE \textit{// Intent Injection: re-state $q$ and $\mathcal{E}$ in observation}
        \STATE $o_t \leftarrow$ retrieved image $I_k$ with intent-injected prompt
        \STATE \textit{// Visual Action Evaluation}
        \STATE \textbf{Evaluation}: guide model to evaluate whether to crop $I_k$
    \ELSIF{$a_t = $ \texttt{crop}}
        \STATE Crop region $c_t$ from the current image $\rightarrow I'$
        \STATE \textit{// Visual Action Correction}
        \STATE \textbf{Correct}: if crop is uninformative, redirect to pre-crop reasoning
        \STATE $o_t \leftarrow I'$ with intent-injected prompt
    \ENDIF
    \STATE Append $(\mathbf{r}_t, o_t)$ to $\mathcal{H}$
\ENDFOR
\STATE \textbf{Force answer}: generate final answer $a$ based on the latest $[\,\mathbf{P};\; \mathcal{E};\; \text{last } W \text{ turns of } \mathcal{H}\,]$
\end{algorithmic}
\end{algorithm}

\paragraph{\textbf{Evidence Collection}.}
Existing agentic RAG systems~\cite{yao2022react, wu2025mmsearch, wang2025vrag} typically process retrieved pages in isolation, making it challenging to jointly reason over evidence scattered across multiple pages. VISOR addresses this limitation by introducing a persistent and structured evidence space $\mathcal{E}$: for each visited page $I_k$, the agent generates up to two text summaries directly into its $\langle$\texttt{think}$\rangle$ block: $e_k^{\text{pre}}$ upon initially inspecting the full page, and $e_k^{\text{post}}$ after examining a localized region via a crop action (which is omitted if no crop occurs).  These summaries are extracted and stored in $\mathcal{E}$:
\begin{equation}
\mathcal{E} = \bigl\{I_k \mapsto (e_k^{\text{pre}},\, e_k^{\text{post}}) \mid k \in \text{retrieved pages}\bigr\}.
\end{equation}
Based on the above design, the agent is explicitly prompted to record \emph{any} potentially useful information from a page, even if that page alone is insufficient to answer the question. This strategy prevents VLMs from prematurely discarding partially relevant content and ensures that fine-grained clues from disparate pages can be effectively synthesized when formulating subsequent search queries or deriving the final answer.

\paragraph{\textbf{Action Space.}}
At each turn, the agent selects one of three actions:
\textbf{(1)~Retrieval}: The agent submits a text query wrapped in \texttt{<search>} tags; the retrieval engine returns the most relevant page (top-1) from $\mathcal{C}$ as the subsequent observation.
\textbf{(2)~Crop}: the agent sends a \texttt{<bbox>} action specifying pixel coordinates to zoom into a specific region of the current page for fine-grained reading.
\textbf{(3)~Answer}: the agent wraps its final response in \texttt{<answer>} tags, terminating the loop.

To minimize unnecessary crop actions, we introduce a \emph{Visual Action Evaluation and Correction} mechanism. Upon receiving a retrieved page, the agent first evaluates whether the globally visible content is already sufficient; if so, no crop is performed. The agent issues a crop action only when a specific region appears to contain relevant content but lacks sufficient clarity or resolution for reading. Furthermore, if a crop yields uninformative results, a \emph{Correction} step is applied: via prompting, the agent's attention is redirected back to its pre-crop reasoning, thereby preventing irrelevant visual content from injecting noise into $\mathcal{E}$.

\paragraph{\textbf{Agent Loop Context.}}
In the visual setting, naively appending every response and observation to the context during long-horizon interactions is prohibitively costly: each retrieved page-level image introduces thousands of tokens, rapidly diluting earlier evidence. To mitigate this, we replace the standard append-only trajectory with a \emph{Dynamic Trajectory} mechanism that dynamically reconstructs the input context at each turn as:
\begin{equation}\label{eq:sliding_window}
\mathbf{C}_t = \bigl[\,\mathbf{P}\;;\;\mathcal{E}_t\;;\;\underbrace{(\mathbf{r}_{t-W+1}, o_{t-W+1}), \ldots, (\mathbf{r}_{t}, o_{t})}_{\text{last } W \text{ turns}}\,\bigr],
\end{equation}
where $\mathbf{P}$ is the initial prompt containing $q$, $\mathcal{E}_t$ is the current evidence space, $\mathbf{r}_i$ and $o_i$ denote the agent's response and the environment observation at turn $i$, and $W$ is the \textbf{sliding window} size that controls how many recent turns are retained. The query $\mathbf{P}$ and $\mathcal{E}_t$ are persistently pinned at the top, ensuring that all cross-page findings accumulated thus far remain visible at every turn. We empirically set $W = 2$ ; thus, when the agent issues a crop action, the context retains the preceding turn (containing the original full-page image and the initial reasoning). This allows the agent to observe the complete \textit{retrieve-then-crop} reasoning chain. While earlier turns are evicted from the raw context window, their distilled evidence remains safely preserved in $\mathcal{E}_t$. This strategy strictly bounds the number of in-context image tokens throughout long-horizon reasoning tasks at inference time.

Furthermore, every system-returned observation—whether triggered by a \texttt{search}, \texttt{crop}, or forced-answer event—incorporates an \emph{Intent Injection} prompt. This prompt explicitly restates the original query $q$ and redirects the model's attention to the evidence space $\mathcal{E}$, reminding the agent to consult its accumulated findings rather than reasoning from scratch. This mechanism ensures the agent remains anchored to its primary objective throughout the trajectory, effectively countering the common tendency of long-horizon reasoning to drift into irrelevant tangents.

\subsection{Training Pipeline}\label{sec:training}

\paragraph{\textbf{Supervised Warm-up via Trajectory Distillation}}
\label{sec:sft}

Inspired by trajectory distillation~\cite{wang2025vrag}, we distill agentic trajectories from Qwen3-VL-235B-A22B~\cite{bai2025qwen3} to construct SFT data. We collect training queries from the training split of SlideVQA~\cite{tanaka2023slidevqa} and apply strict quality filtering——particularly on retrieval steps—discarding trajectories that exhibit redundant or uninformative searches. This yields about 2.5K curated trajectories (see supplementary material for details). 
This SFT stage endows VLM with two complementary capabilities.
First, it learns the foundational agentic skills required by our framework: strict adherence to structured output formats (\texttt{<think>}, \texttt{<search>}, \texttt{<bbox>}, \texttt{<answer>} tags) and precise formulation of retrieval intents.
Second, it acquires a verification search strategy: after gathering sufficient evidence, the agent executes one final search query to confirm that no critical visual or textual information has been overlooked before generating its answer.
This deliberate verification step contrasts sharply with prior agentic VRAG systems~\cite{wu2025mmsearch, wang2025vrag}, which typically suppress additional searches to mitigate context bloat. In contrast, our VISOR framework dynamically reconstructs and compresses trajectories during inference, enabling it to reward—rather than penalize—such verification behavior.

We emphasize that the SFT data consists of conventional long-horizon trajectories, without evidence-space abstraction or sliding-window context restructuring. 
During training, observation tokens are masked, ensuring that gradients propagate exclusively through the model's own responses.
Introducing the dynamic context format at the SFT stage would unnecessarily complicate data curation and is orthogonal to its objective: SFT is designed solely to learn basic agentic behaviors and format compliance, while leaving context-efficient reasoning to the subsequent RL stage.

\paragraph{\textbf{Reinforcement Learning}}
\label{sec:rl}

While SFT enables the agent to imitate expert trajectories, imitation alone provides only a weak and indirect signal for determining \emph{when} to terminate retrieval: the teacher's stopping behavior lacks explicit optimality guarantees and is not reinforced by reward feedback.
The RL stage addresses this limitation. Through self-generated rollouts in the retrieval environment, the agent receives dense, task-aligned rewards, allowing it to discover and refine retrieval strategies that surpass what pure imitation can convey, including precise stopping criteria and adaptive search depth. We optimize the policy using GRPO~\cite{shao2024deepseekmath} under the \textit{dynamic trajectory} described above.

Two designs address the unique challenges of multi-turn agentic training. First, \emph{observation masking}: environment-provided observations, such as retrieved images or cropped regions, are external to the agent's policy. To ensure proper credit assignment, we exclude these tokens from the policy gradient. And only the agent's own generated tokens \texttt{<think>}/\texttt{<action>} contribute to parameter updates. Second, \emph{credit assignment under dynamic context}: the sliding window is applied at generation time only. Specifically, each forward pass uses the dynamically reconstructed context for token prediction, whereas the loss is computed over the entire uncompressed response sequence. This ensures that both reward calculation and gradient backpropagation operate on the complete, untruncated trajectory, enabling accurate long-horizon credit assignment despite context compression during generation.

The total reward $r$ first checks format validity, then combines retrieval and answer quality:
\begin{equation}
r = (r_\text{ans} + r_\text{ret})\bigl(1-\mathbb{I}_{\text{format}}\bigr)-1 \cdot \mathbb{I}_{\text{format}}.
\end{equation}

\noindent\textbf{Answer Reward ($r_\text{ans}$).}~$r_\text{ans}$ is gated on whether the retrieval process succeeded:
\begin{equation}
r_\text{ans} =
\begin{cases}
\text{LLM-judge}(\hat{a},\, a^*) \in \{0, 1\} & \text{retrieval complete} \\
\text{honesty-judge}(\hat{a}) \in \{0, 0.2\} & \text{retrieval incomplete}
\end{cases}
\end{equation}
When retrieval is complete, an LLM judge (Qwen-Max-Latest~\cite{bai2025qwen3}) compares the predicted answer $\hat{a}$ against the reference $a^*$ and returns a binary correctness score. When retrieval is incomplete, we instead apply an \emph{epistemic honesty} check: the agent receives partial reward (0.2) for explicitly acknowledging insufficient information rather than fabricating a response. We gate these two evaluations on retrieval completeness to prevent reward hacking—without this gate, an agent that fails to retrieve the necessary pages could still earn high answer credit by producing fluent-sounding but fabricated responses that happen to fool the judge.

\noindent\textbf{Retrieval Reward ($r_\text{ret}$).}~Let $t^*$ be the turn at which all reference pages are first retrieved and $T$ the total number of search turns. We define $\Delta = T - t^*$ as the number of extra search turns after all necessary evidence has been gathered:
\begin{equation}
r_\text{ret} =
\begin{cases}
-1.0 & \text{incomplete retrieval (missing evidence pages)} \\
-0.5 & \Delta = 0\;\text{(no verification search)} \\
\phantom{-}0.0 & \Delta = 1\;\text{(exactly one verification, optimal)} \\
-0.1{\times}\Delta & \Delta \geq 2\;\text{(over-searching)}
\end{cases}
\end{equation}
Our reward shaping explicitly encourages exactly one verification search in each trajectory: stopping immediately after full retrieval ($\Delta=0$) is unrewarded, risking missed evidence, while excessive searching ($\Delta\geq2$) is penalized.
Failure to retrieve all necessary pages incurs the harshest penalty, as missing evidence is unrecoverable by reasoning alone.

\noindent\textbf{Format Penalty ($\mathbb{I}_{\text{format}}$).}~Besides the above two reward scores, we also introduce an output penalty score $\mathbb{I}_{\text{format}}$ for the whole trajectory format inspired by~\citep{guo2025deepseek}. $\mathbb{I}_{\text{format}}$ is set to 1 if the trajectory format is invalid (invokes no \texttt{search} action before the \texttt{answer} action), and 0 otherwise.

\begin{table*}[htbp]
\centering
\caption{Main results on SlideVQA, ViDoSeek, and MMLongBench. We report accuracy (\%). $\dagger$ denotes multi-agent architectures. $\ddagger$ denotes fine-tuned models. $\star$ denotes results taken from published papers under the same or comparable experimental settings. The best result in each column is \textbf{bolded} and the second-best is \underline{underlined}.}
\label{tab:main}
\resizebox{0.9\textwidth}{!}{%
\begin{tabular}{lccc|ccc|cccccc}
\toprule
\multirow{2}{*}{\textbf{Method}} &
  \multicolumn{3}{c}{\textbf{SlideVQA}} &
  \multicolumn{3}{c}{\textbf{ViDoSeek}} &
  \multicolumn{6}{c}{\textbf{MMLongBench}} \\
\cmidrule(lr){2-4}\cmidrule(lr){5-7}\cmidrule(lr){8-13}
 & Single-hop & Multi-hop & \textbf{Overall} & Extraction & Logic & \textbf{Overall} & Text & Table & Chart & Figure & Layout & \textbf{Overall} \\
\midrule
\rowcolor{gray!15}
\multicolumn{13}{c}{\textit{Qwen2.5-VL-7B}} \\
\midrule
Vanilla RAG$^\star$~\cite{faysse2024colpali}      & 29.10 & 17.40 & 26.10 & 26.40 & 41.30 & 32.88 & 13.10 & 14.70 & 15.90 & 4.30  & 7.60  & --    \\
ReAct$^\star$~\cite{yao2022react}                 & 34.80 & 20.40 & 31.11 & 27.50 & 42.10 & 33.85 & 10.10 & 12.40 & 10.20 & 6.20  & 7.10  & --    \\
ViDoRAG$^\dagger$~\cite{wang2025vidorag}          & 72.15 & 39.86 & 63.88 & 66.05 & 72.83 & 69.00 & 24.40 & 23.96 & 21.91 & 24.14 & 20.34 & 25.50 \\
M3RAG$^{\dagger\star}$~\cite{du2026m3rag}          & --    & --    & 65.82 & --    & --    & 69.36 & --    & --    & --    & --    & --    & --    \\
Search-R1-VL$^{\ddagger\star}$~\cite{jin2025search}  & 48.30 & 42.30 & 46.76 & 40.50 & 50.30 & 44.77 & 19.90 & 13.40 & 12.90 & 11.40 & 10.20 & --    \\
VRAG-RL$^{\ddagger\star}$~\cite{wang2025vrag}        & 69.30 & 43.10 & 62.59 & 60.60 & \underline{74.80} & 66.78 & 26.10 & \underline{26.30} & \underline{24.80} & \underline{25.90} & \underline{21.20} & --    \\
MMSearch-R1$^\ddagger$~\cite{wu2025mmsearch}          & 52.06 & 40.21 & 49.03 & 55.97 & 59.56 & 57.53 & 16.84 & 17.97 & 19.10 & 18.28 & 11.86 & 18.42 \\
EVisRAG$^\ddagger$~\cite{sun2025visrag}               & \underline{78.21} & 42.32 & \underline{69.09} & \underline{67.75} & 72.43 & \underline{69.79} & \underline{26.80} & \textbf{27.65} & 24.72 & 22.41 & 19.49 & \underline{27.98} \\
R1-Router$^{\ddagger}$~\cite{peng2025learning}   & 69.66 & \underline{45.33} & 63.43 & 64.19 & 68.61 & 66.11 & 26.46 & 23.50 & 22.47 & \textbf{28.28} & 16.95 & 26.92 \\
\textbf{VISOR}$^\ddagger$                          & \textbf{78.82}  & \textbf{53.62} & \textbf{72.37}  & \textbf{73.49}  & \textbf{76.66} & \textbf{74.87}  & \textbf{27.49} & 23.96  & \textbf{27.53}  & 23.79  & \textbf{22.88}  & \textbf{28.45}  \\
\midrule
\rowcolor{gray!15}
\multicolumn{13}{c}{\textit{Qwen2.5-VL-3B}} \\
\midrule
Vanilla RAG$^\star$~\cite{faysse2024colpali}      & 19.40 & 12.20 & 17.56 & 10.10 & 17.30 & 13.23 & 2.20  & 4.10  & 5.20  & 4.70  & 4.30  & --    \\
ReAct$^\star$~\cite{yao2022react}                 & 15.70 & 10.90 & 14.47 & 6.70  & 14.20 & 9.96  & 2.70  & 3.60  & 3.40  & 3.10  & 5.10  & --    \\
ViDoRAG$^\dagger$~\cite{wang2025vidorag}          & 41.44 & 19.93 & 35.94 & 31.32 & 38.23 & 34.33 & 7.90  & 6.91  & 5.06  & 8.97  & 7.63  & 8.50  \\
Search-R1-VL$^{\ddagger\star}$~\cite{jin2025search}  & 26.30 & 20.10 & 24.71 & 20.10 & 29.80 & 24.32 & 8.50  & 7.80  & 7.90  & 9.30  & 7.60  & --    \\
VRAG-RL$^{\ddagger\star}$~\cite{wang2025vrag}        & 65.30 & 38.60 & 58.45 & 63.10 & \textbf{73.80} & 67.76 & 22.70 & 16.10 & 21.90 & 21.40 & 19.50 & --    \\
EVisRAG$^\ddagger$~\cite{sun2025visrag}               & \textbf{75.42} & \underline{47.70} & \underline{68.35} & \underline{66.05} & \underline{72.43} & \underline{68.82} & \textbf{27.14} & \textbf{28.64} & \underline{25.13} & \underline{24.83} & 17.80 & \textbf{28.34} \\
R1-Router$^{\ddagger}$~\cite{peng2025learning}   & 64.93 & 42.15 & 59.10 & 62.64 & 69.42 & 65.59 & \underline{26.80} & 23.50 & 21.91 & \textbf{25.17} & \underline{21.19} & 25.74 \\
\textbf{VISOR}$^\ddagger$                          & \underline{74.58}  & \textbf{50.79} & \textbf{68.49}  & \textbf{67.75}  & 70.62 & \textbf{69.00}  & \underline{26.80}  & \underline{23.96}  & \textbf{26.97}  & 23.45  & \textbf{22.03}  & \underline{27.86}  \\
\bottomrule
\end{tabular}%
}
\end{table*}

\section{Experiments}
\subsection{Experimental Settings}

\paragraph{\textbf{Datasets and Metric}}

Following VRAG-RL~\cite{wang2025vrag}, we evaluate our method on three challenging and visually rich benchmarks: \textbf{ViDoSeek}~\cite{wang2025vidorag}, \textbf{SlideVQA}~\cite{tanaka2023slidevqa}, and \textbf{MMLongBench}~\cite{ma2024mmlongbench}. For each benchmark, all images of document pages are pooled into a unified retrieval corpus, and the model retrieves relevant images from this corpus to answer each question. For fine-tuning, we use training data derived from the SlideVQA training split, comprising 2,500 samples for supervised fine-tuning (SFT) and 800 samples for reinforcement learning (RL). To evaluate the performance, we employ \texttt{Qwen-max-latest} as the judge model to assign a binary score (0 or 1) to each response by comparing it against the reference answer, and we report the mean score as the final accuracy. Further details are provided in the supplementary material.

\paragraph{\textbf{Baselines}}
For vanilla (non-fine-tuned) baselines, we compare against Vanilla RAG~\cite{faysse2024colpali} and ReAct~\cite{yao2022react}, as well as two multi-agent architectures: ViDoRAG~\cite{wang2025vidorag} and M3RAG~\cite{du2026m3rag}. For fine-tuned baselines, we compare with Search-R1-VL~\cite{jin2025search}, VRAG-RL~\cite{wang2025vrag}, MMSearch-R1~\cite{wu2025mmsearch}, R1-Router~\cite{peng2025learning}, and EVisRAG~\cite{sun2025visrag}. All methods use ColQwen2.5-v0.1~\cite{faysse2024colpali} as the shared retrieval backbone, ensuring a fair comparison. Implementation details of all baselines are provided in the supplementary material.

\subsection{Main Results}

As shown in Table~\ref{tab:main}, VISOR achieves state-of-the-art performance across all three benchmarks and all backbone sizes.
Specifically, on SlideVQA, the improvement is most pronounced on multi-hop questions, where our VISOR scores 53.62\%, substantially outperforming VRAG-RL (43.10\%) and EVisRAG (42.32\%). This demonstrates that our Evidence Space effectively supports cross-page reasoning. In contrast, performance on single-hop questions remains comparable to the best baseline (78.82\%), indicating that VISOR introduces no overhead on simpler queries.
On ViDoSeek, gains are markedly larger for Extraction task (+12.89\% over VRAG-RL) than for Logic task (+1.86\%). This suggests that our Evidence Space is particularly beneficial for retrieving and aggregating factual content across pages, whereas logical inference—once relevant pages are identified—is already well addressed by existing RL-trained reasoning modules.
On MMLongBench, VISOR leads on Chart and Layout subtasks (+2.81\% and +3.39\% over EVisRAG, respectively), highlighting the importance of our selective crop-and-zoom mechanism for fine-grained visual understanding.

With the smaller \textbf{Qwen2.5-VL-3B} backbone, VISOR achieves 68.49\% on SlideVQA, 69.00\% on ViDoSeek, and 27.86\% on MMLongBench, outperforming EVisRAG on the first two benchmarks.
The slight performance gap on MMLongBench (vs. 28.34\% from EVisRAG) stems primarily from differing training objectives: EVisRAG is fine-tuned on large-scale multi-image understanding data, granting it stronger multi-image comprehension capability, while VISOR focuses on agentic retrieval and long-horizon reasoning.
On visually richer content, the reduced capacity of the 3B backbone amplifies this architectural divergence.
Notably, multi-agent methods suffer severe degradation at this scale, e.g., ViDoRAG performance drops sharply from 69.00\% to 34.33\% on ViDoSeek. In contrast, fine-tuned single-agent methods exhibit only marginal decline (e.g., EVisRAG: 69.79\%$\to$68.82\%), underscoring that end-to-end optimization becomes increasingly critical as model capacity decreases.

\subsection{Ablation Study}
\label{ablation}
\begin{table}[b]
\centering
\caption{Ablation study of VISOR on Qwen2.5-VL-7B on SlideVQA and ViDoSeek. We report overall accuracy (\%). II = Intent Injection, ES = Evidence Space with Visual Action Evaluation and Correction, SW = Sliding Window.}
\label{tab:ablation}
\resizebox{0.9\columnwidth}{!}{%
\begin{tabular}{lcccc}
\toprule
\multirow{2}{*}{\textbf{Method}} & \multicolumn{2}{c}{\textbf{ViDoSeek}} & \multicolumn{2}{c}{\textbf{SlideVQA}} \\
\cmidrule(lr){2-3} \cmidrule(lr){4-5}
 & \textbf{Vanilla} & \textbf{Fine-tuned} & \textbf{Vanilla} & \textbf{Fine-tuned} \\
\midrule
\textbf{VISOR} (Full)       & \textbf{57.88} & \textbf{74.87} & \textbf{54.00} & \textbf{72.37} \\
\midrule
w/o II                      & 49.04           & 69.00           & 43.52          & 69.07          \\
w/o ES                      & 39.58           & 47.81           & 39.10          & 51.33          \\
w/o SW                      & 53.94           & 70.93           & 47.47          & 67.54          \\
w/o ES \& SW                & 42.11           & 68.48           & 40.81          & 64.61          \\
w/o II \& ES \& SW          & 37.57         & 66.78         & 36.98          & 62.59          \\
\bottomrule
\end{tabular}%
}
\end{table}

As shown in Table~\ref{tab:ablation}, we ablate VISOR by decomposing three key components: Intent Injection (II), Evidence Space with Visual Action Evaluation and Correction (ES), and Sliding Window (SW).
All ablation studies are conducted using Qwen2.5-VL-7B backbone on both ViDoSeek and SlideVQA.
The results demonstrate that the removal of any single component leads to consistent performance degradation across both datasets and evaluation settings, underscoring the necessity and complementary nature of each module.

\emph{\textbf{Detailed Analysis of Evidence Space (ES) and Sliding Window (SW).}}
(1) The removal of ES incurs the largest accuracy drop across both benchmarks and settings. This indicates that parameter updates alone cannot compensate for the loss of evidence distillation capability.
(2) Eliminating SW also consistently degrades performance, as the model is forced to process an unbounded sequence of visual tokens without any context budgeting strategy, quickly exceeding its effective receptive capacity.
(3) Most revealing, however, is their interaction: when ES is absent, further removing SW partially recovers performance. This counterintuitive result demonstrates that SW is only beneficial when paired with ES. Without ES to extract and preserve critical evidence from discarded turns, SW truncation becomes purely destructive, discarding raw context with no compensatory signal. Conversely, without SW, the model retains all historical turns but rapidly exhausts its context window, leading to inefficiency and degradation.
(4) Together, these findings reveal a tight coupling between the two components: SW enables efficient, bounded-context processing by compressing the interaction trajectory, while ES ensures that semantically valuable information lost during compression is retained through structured evidence distillation. Only when jointly deployed do ES and SW effectively tackle the central challenge of long-horizon visual retrieval; in isolation, each component proves either inadequate or actively detrimental.

\emph{\textbf{Detailed Analysis of Intent Injection (II).}}
In the absence of II, the model lacks explicit query re-anchoring at each interaction turn. This leads to a noticeable performance drop on both benchmarks. However, the gap narrows substantially after fine-tuning, suggesting that the model can partially remedy query-aware behavior through training. 
Thus, II primarily serves as an architectural inductive bias that guides untrained models toward query-aligned reasoning, while still offering a consistent (though reduced) benefit in fine-tuned settings.

In summary, removing all three components (II, ES, and SW) reduces VISOR to a vanilla agentic baseline. The persistent performance gap across both benchmarks underscores that these modules jointly mitigate two core challenges in long-horizon visual retrieval: sparse evidence signals and search drift. Their synergy—not just their individual contributions—is key to VISOR's effectiveness.
\subsection{Analysis}

\paragraph{\textbf{Retrieval Behavior Analysis}}
Understanding the limitations of existing retrieval logic is key to motivating our design. 
In particular, methods like VRAG-RL illustrate a fundamental tension between retrieval completeness and context length, a tradeoff that our VISOR framework explicitly seeks to resolve.

\begin{figure}[t]
\centering
\includegraphics[width=0.7\columnwidth]{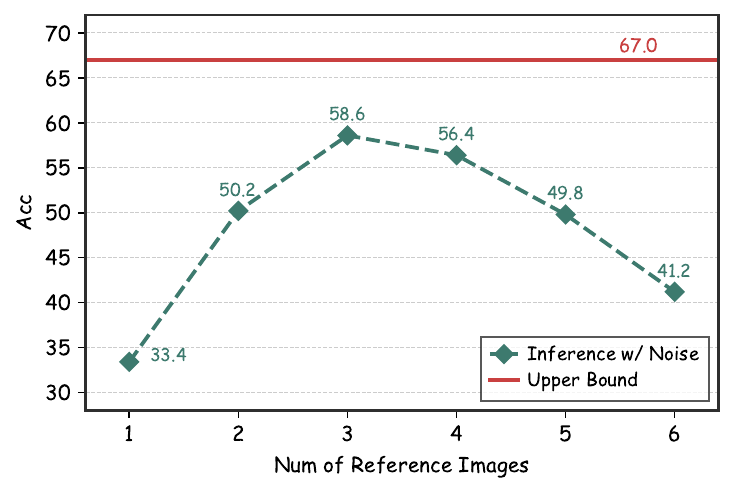}
\caption{Accuracy of Qwen2.5-VL-7B on SlideVQA (500 samples) as a function of the number of retrieved reference images. The upper bound (67.0\%) is the accuracy when correct reference images are directly provided to the model.}
\label{fig:retrieval_behavior}
\end{figure}

Figure~\ref{fig:retrieval_behavior} analyzes Qwen2.5-VL-7B on 500 SlideVQA samples, plotting accuracy as a function of the number of retrieved reference images. 
At low retrieval counts, the model often retrieves only visually similar but non-informative pages or misses critical evidence required for multi-hop reasoning, forcing it to answer with incomplete information. 
As more images are retrieved, however, irrelevant content accumulates, inducing semantic drift that obscures earlier findings and ultimately degrades performance. 
The upper-bound curve, representing accuracy when ground-truth reference images are directly provided, confirms that \textit{completeness} of retrieved content, and semantic \textit{drift} caused by long context, remains a primary bottleneck.

As shown in Table~\ref{tab:retrieval_completeness}, this tradeoff between completeness and drift manifests differently across methods. 
On the \emph{completeness} axis, methods that perform retrieval only once at the beginning (e.g., EVisRAG) are structurally constrained: a fixed top-$k$ retrieval cannot recover missing evidence if relevant pages lie outside the initial set, especially for multi-hop questions. ViDoRAG mitigates this by retrieving up to 10 pages by default, achieving the highest completeness, and yet this brute-force expansion introduces substantial noise into the context.
On the \emph{drift} axis, methods like VRAG-RL iteratively gather evidence over multiple interaction steps, improving coverage beyond a single retrieval pass; yet without explicit evidence management, each new image appends raw visual tokens to an ever-growing context, shifting the failure mode from missing evidence to accumulated irrelevance.
Our VISOR addresses this dilemma at the architectural level. By distilling verified evidence into a compact, structured representation and enforcing a bounded context window, VISOR enables verification-driven search without uncontrolled noise accumulation, simultaneously achieving high retrieval completeness and sustained contextual coherence.

\begin{table}[htbp]
\centering
\caption{Retrieval completeness (\%) and average number of retrieved images per trajectory on SlideVQA (2,215 samples). Completeness = fraction of questions where all ground-truth evidence pages are retrieved.}
\label{tab:retrieval_completeness}
\small
\begin{tabular}{lcc}
\toprule
\textbf{Method} & \textbf{Completeness (\%)} & \textbf{Avg.\ Retrieved} \\
\midrule
EVisRAG~\cite{sun2025visrag} & 80.2 & 3 (fixed) \\
VRAG-RL~\cite{wang2025vrag}        & 76.8 & 1.78 \\
ViDoRAG~\cite{wang2025vidorag}        & 91.3 & 10 (fixed) \\
\textbf{VISOR}   & \textbf{84.2} & \textbf{2.34} \\
\bottomrule
\end{tabular}
\end{table}

\begin{figure}[htbp]
\centering
\includegraphics[width=0.75\columnwidth]{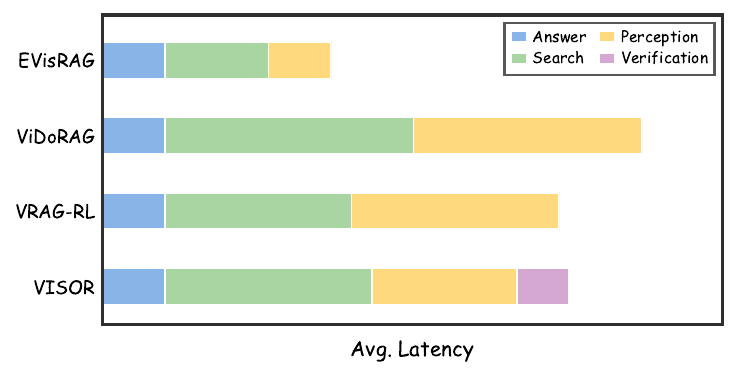}
\caption{Breakdown of average per-sample inference latency on ViDoSeek. VISOR introduces a verification step yet remains faster than the multi-agent baseline ViDoRAG and competitive with VRAG-RL.}
\label{fig:time_efficiency}
\end{figure}

\paragraph{\textbf{Time Efficiency}}
As shown in Figure~\ref{fig:time_efficiency}, VISOR is naturally slower than \textbf{EVisRAG}, which performs a single fixed top-$k$ retrieval and one-shot answer, due to its iterative multi-turn interaction. Compared to \textbf{VRAG-RL}, another iterative method, VISOR introduces an additional verification turn yet remains competitive in overall latency. This is because the Sliding Window prevents context length from growing unboundedly across turns, significantly accelerating per-turn inference, while Visual Action Evaluation reduces unnecessary crop-and-zoom operations by issuing bounding-box actions only when fine-grained inspection is genuinely needed. Compared to \textbf{ViDoRAG}, VISOR is notably faster despite achieving higher accuracy. ViDoRAG retrieves up to 10 pages per query and subsequently runs repeated actor-critic cycles over the retrieved content through multiple specialized agents, incurring substantial VLM inference overhead. VISOR avoids this by adaptively retrieving only the pages needed per query, keeping the average retrieved count low while maintaining high completeness.

\paragraph{\textbf{Effect of Training.}}
Table~\ref{tab:training_effect} examines how each training stage shapes the model's behavior on ViDoSeek.
The \textit{finish rate} measures the proportion of episodes in which the model produces a non-empty final answer, reflecting whether the agent successfully completes the task within the allocated number of steps. The \textit{invalid action rate} measures the proportion of episodes in which at least one action violates the required format constraints (e.g., missing mandatory \texttt{<think>} reasoning or issuing malformed tool calls); upon detecting such an action, the framework immediately returns a system-level error message prompting the model to correct its behavior before proceeding.

\begin{table}[htbp]
\centering
\caption{Effect of training stage on the model.}
\label{tab:training_effect}
\resizebox{0.8\columnwidth}{!}{%
\begin{tabular}{lccc}
\toprule
\textbf{Method} & \textbf{Invalid Action Rate ↓} & \textbf{Finish Rate ↑} & \textbf{Acc ↑} \\
\midrule
Vanilla                  & 0.61 & 74.08 & 57.88 \\
SFT                      & 0.53 & 79.42 & 63.57 \\
\textbf{SFT+RL}  & \textbf{0.09} & \textbf{96.23} & \textbf{74.87} \\
\bottomrule
\end{tabular}%
}
\end{table}

The vanilla model already benefits from VISOR's framework structure, but frequently produces invalid actions—failing to follow the required output format or issuing hallucinated crop coordinates—and often terminates prematurely without retrieving sufficient evidence.
SFT on long-horizon trajectories substantially reduces the invalid action rate and improves finish rate, demonstrating that models trained in the standard trajectory format can be directly deployed within VISOR's framework without architectural modification.
RL further refines decision-making at each step: the model learns when to search, when to crop, and when to answer, leading to the lowest invalid rate, highest finish rate, and best accuracy.
These results suggest that VISOR's framework is broadly compatible with existing training paradigms and does not require specialized trajectory formats to benefit from.

\paragraph{\textbf{Visual Action Usage Analysis.}}
Crop-and-zoom enables fine-grained perception of information-dense regions, but indiscriminate cropping introduces visual noise and wastes interaction turns. Without an explicit gating mechanism, models tend to over-apply crop actions—supplementary material shows a representative case where a crop is applied to an already-legible region, gaining nothing while consuming an extra turn.
VISOR's \emph{Visual Action Evaluation} mechanism explicitly prompts the model to assess whether cropping is necessary before acting, directly suppressing such redundant operations. As a result, VISOR's bbox usage rate on ViDoSeek is only \textbf{6.65\%}, compared to \textbf{82.92\%} in VRAG-RL where no such gate exists. The crops that are issued under VISOR are genuinely informative, as reflected by the gains on Chart and Layout in Table~\ref{tab:main}.

\paragraph{\textbf{Reliability of Model-as-Judge.}}
We use an LLM judge (Qwen-Max-Latest) both as the answer reward signal during RL training and as the evaluation metric at test time. Compared to rule-based alternatives such as exact match, a model-based judge handles the linguistic variability inherent in open-ended visual question answering—where correct answers may be expressed in multiple valid forms—without over-penalizing semantically equivalent responses. Compared to recall-based soft matching, it is less susceptible to reward hacking through repetitive or loosely overlapping outputs. We verify the reliability of this judge by measuring its agreement with human annotations on a held-out subset; results (in the supplementary material) confirm high consistency, supporting its use as both a training signal and an evaluation criterion.

\paragraph{\textbf{Case Study}}
We present representative success and failure trajectories of VISOR in the supplementary material. In the success case, VISOR handles a multi-hop question spanning two separate pages: it retrieves each relevant page in turn, distills findings into the Evidence Space, and synthesizes a correct answer from the accumulated evidence—a capability that single-pass retrieval and page-isolated reasoning cannot provide. In the failure case, VISOR successfully retrieves the correct page yet produces an incorrect answer due to misinterpretation of a complex structural diagram. This suggests that while VISOR's retrieval and evidence management are effective, visual understanding of intricate visual structures remains a bottleneck, pointing to a clear direction for future improvement.

\section{Conclusion}

We presented \textbf{VISOR}, a framework for agentic VRAG that tackles visual evidence sparsity and search drift in long-horizon document understanding. Through deliberate architectural design, VISOR achieves structured evidence management within an end-to-end trainable loop, without sacrificing retrieval flexibility or reasoning coherence. Experiments confirm state-of-the-art results on SlideVQA and ViDoSeek, and ablations show that the Evidence Space addresses a structural bottleneck that RL training alone cannot resolve. We hope VISOR offers a practical blueprint for building capable, efficient agentic VLMs over visually rich corpora.

\begin{acks}
This work is supported by the National Natural Science Foundation of China under Grants 62476188.  
\end{acks}


\bibliographystyle{ACM-Reference-Format}

\bibliography{main}

\clearpage
\appendix

\section{Search Engine and Crop-and-Zoom Tool}\label{app:tools}

\paragraph{Search Engine.}
We use ColQwen2.5-v0.1~\cite{faysse2024colpali} as our retrieval backbone.
Each document page is pre-encoded into patch-level multi-vector embeddings offline and stored for fast lookup.
At inference time, the agent's text query is encoded by the same model and ranked against all page embeddings via the MaxSim operator, returning the top-$k$ candidate pages.
To avoid redundancy, the environment maintains a per-trajectory retrieval history: from the ranked list, the first page not previously shown to the agent is selected as the observation.
If all top-$k$ candidates have already been retrieved in earlier turns, the agent receives a prompt indicating that no new pages are available and is directed to produce a final answer.

\paragraph{Crop-and-Zoom Tool.}
When the agent emits a \texttt{<bbox>[x1, y1, x2, y2]</bbox>} action, the pixel coordinates—expressed in the VLM's displayed image space—are linearly mapped back to the original high-resolution page.
Before cropping, the bounding box is expanded by a fixed margin of 28 pixels on each side to preserve contextual content near the boundary, and then clamped to the image extent to prevent out-of-bound access.
The resulting region is cropped and resized to a standard resolution, enabling the agent to read fine-grained content (\emph{e.g.}, dense tables or small charts) that is difficult to perceive at full-page scale.
Invalid coordinates result in an error message prompting the model to retry.

\section{SFT Data Construction}\label{app:sft_data}

\paragraph{Raw Data Collection.}
Following VRAG-RL~\cite{wang2025vrag}, we construct SFT data by collecting expert trajectories via prompt-based distillation.
Training queries are drawn from the training split of SlideVQA~\cite{tanaka2023slidevqa}.
For each query, we prompt the teacher model to generate a complete agentic trajectory—comprising interleaved search queries, crop-and-zoom actions, and reasoning steps—using the same ReAct-style format defined in our system prompt.
In contrast to VRAG-RL, which employs separate models for trajectory generation and grounding, we use Qwen3-VL-235B-A22B~\cite{bai2025qwen3} uniformly for all data collection steps.

\paragraph{Data Filtering.}
We apply a series of quality filters to retain only high-quality trajectories.
First, we enforce structural validity of crop actions: any trajectory where a \texttt{<bbox>} action is not immediately preceded by a retrieval step is discarded, as such actions lack grounding in the retrieved content.
We additionally remove trajectories containing trivial full-image crops (coordinates $[0, 0, 1000, 1000]$), which provide no additional information over the original page view.
Second, we apply a retrieval completeness criterion: a trajectory is retained only if all ground-truth reference pages appear among its retrieved images, ensuring that the SFT signal is grounded in sufficient evidence.
Finally, trajectories with more than 10 retrieval steps are discarded to avoid excessively long context sequences.
These filters together yield a candidate pool of trajectories, which are further filtered by answer correctness: only trajectories where the final answer matches the reference answer are retained.
This yields approximately 2.5K high-quality training trajectories.

\paragraph{Data Augmentation.}
After filtering, we apply two targeted modifications to each trajectory.
First, all search queries within a trajectory are replaced with the original question text.
The teacher model often generates paraphrased or reformulated queries that may omit key details from the original question due to intent drift; the original question, by contrast, contains all critical information needed for retrieval, ensuring every search step stays anchored to the user's intent.
Second, we insert a mandatory verification round at the end of each trajectory.
Specifically, the final think in the last assistant turn is appended with a verification intent statement, and the original \texttt{<answer>} action is replaced by a new \texttt{<search>} action.
A previously unseen page from the same document is then randomly sampled as the verification image.
We use Qwen3-VL-235B-A22B~\cite{bai2025qwen3} to analyze the new image in the context of the already-found answer, producing a verification reasoning text.
A final assistant turn is then appended, containing the verification reasoning followed by the original answer.
This augmentation teaches the model to perform one last confirmatory retrieval before committing to an answer, directly corresponding to the verification behavior elicited at inference time by our \emph{Verification Hint} prompt.

\section{Training Hyperparameters}\label{app:hyperparams}

The detailed hyperparameters used during SFT and RL training are listed in Table~\ref{tab:hyper_sft} and Table~\ref{tab:hyper_rl}, respectively.
All experiments reported in the main paper are conducted with Qwen2.5-VL-7B as the backbone on a single node of 8 NVIDIA A800 GPUs.
While VISOR is model-agnostic and can be applied to VLMs of different scales, practitioners may find it beneficial to adjust hyperparameters such as learning rate and context length proportionally when scaling to larger or smaller models.

\begin{table}[h]
\centering
\caption{Key hyperparameters for SFT.}
\label{tab:hyper_sft}
\begin{tabular}{lc}
\hline
\textbf{Name} & \textbf{Value} \\
\hline
Finetuning type              & Full    \\
Freeze vision tower          & True    \\
Freeze multi-modal projector & True    \\
Freeze language model        & False   \\
Cutoff length                & 16384   \\
Epochs                       & 3       \\
Batch size                   & 16      \\
Gradient accumulation steps  & 2       \\
Learning rate                & 1.0e-5  \\
LR scheduler type            & cosine  \\
Warmup ratio                 & 0.1     \\
\hline
\end{tabular}
\end{table}

\begin{table}[h]
\centering
\caption{Key hyperparameters for RL.}
\label{tab:hyper_rl}
\begin{tabular}{lc}
\hline
\textbf{Name} & \textbf{Value} \\
\hline
Number of agent groups       & 5       \\
Warmup steps ratio           & 0.285   \\
Train batch size             & 8       \\
Mini batch size per GPU      & 1       \\
Micro batch size per GPU     & 1       \\
Learning rate (Actor)        & 1.0e-6  \\
KL loss coefficient          & 0.01    \\
Tensor model parallel size   & 2       \\
Max prompt length            & 8192    \\
Max response length          & 2048    \\
Max turns                    & 10      \\
Total steps                  & 100     \\
GPU memory utilization       & 0.4     \\
\hline
\end{tabular}
\end{table}

\section{Evaluation Details}\label{app:eval_details}

\begin{figure*}[htbp]
\centering
\includegraphics[width=0.9\textwidth]{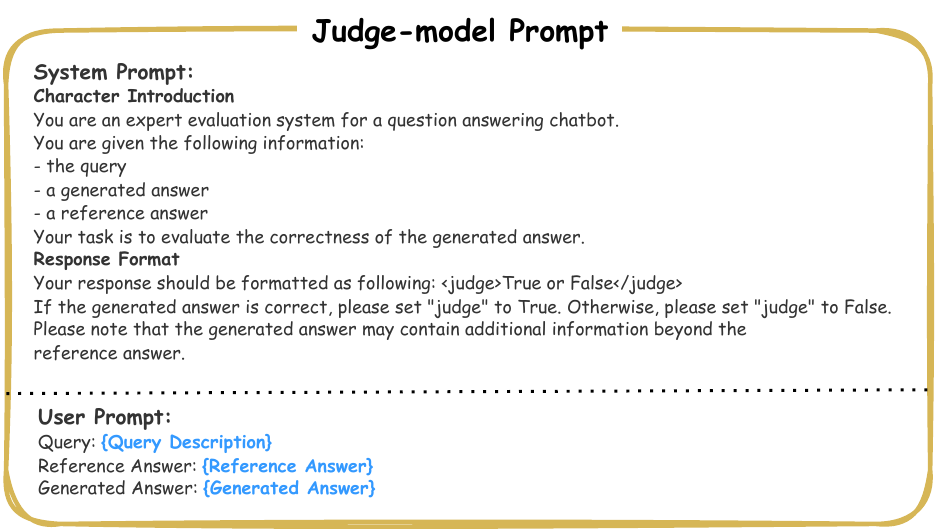}
\caption{The prompt template used for LLM-as-Judge evaluation.}
\label{fig:judge_prompt}
\end{figure*}

\paragraph{Judge Prompt.}
Following VRAG-RL~\cite{wang2025vrag}, we use \texttt{Qwen-max-latest} as the LLM judge to evaluate model responses.
Given the question, reference answer, and model prediction, the judge assigns a binary score (0 or 1) indicating whether the prediction is semantically correct.
The exact prompt used is shown in Figure~\ref{fig:judge_prompt}.

\paragraph{Reliability of Model-as-Judge.}
The reliability of \texttt{Qwen-max-latest} as an LLM judge has been validated in VRAG-RL~\cite{wang2025vrag}.
To further corroborate this, we additionally evaluate our results on ViDoSeek using \texttt{DeepSeek-V3.2}~\cite{liu2024deepseek} as an alternative judge.
As shown in Table~\ref{tab:judge_compare}, the two judges yield highly consistent scores overall, with minor differences at the subtask level: the DeepSeek judge assigns slightly lower scores on Extraction and slightly higher scores on Logic.
The small gap confirms that our reported results are not sensitive to the choice of judge model.

\begin{table}[h]
\centering
\caption{Comparison of evaluation scores on ViDoSeek under two judge models.}
\label{tab:judge_compare}
\begin{tabular}{lccc}
\hline
\textbf{Judge} & \textbf{Extraction} & \textbf{Logic} & \textbf{Overall} \\
\hline
Qwen-max-latest  & 73.49 & 76.66 & 74.87 \\
DeepSeek-V3.2      & 71.47 & 78.67 & 74.61 \\
\hline
\end{tabular}
\end{table}

\section{Dataset Information}\label{app:datasets}

We evaluate VISOR on three visually rich document benchmarks.

\paragraph{SlideVQA.}
SlideVQA~\cite{tanaka2023slidevqa} is a VQA benchmark built on presentation slides, covering a wide range of real-world slide decks across diverse topics.
Questions are split into \textbf{Single-hop} (1,648) and \textbf{Multi-hop} (567) subsets, where multi-hop questions require aggregating evidence from multiple slides within the same deck.
Our evaluation uses the full test split of 2,215 questions.

\paragraph{ViDoSeek.}
ViDoSeek~\cite{wang2025vidorag} is a benchmark targeting retrieval augmented question answering over large-scale visually rich corpora, with approximately 6,000 document page images spanning text, tables, charts, and figures.
The test set of 1,142 questions is divided into two non-overlapping subsets by question type: \textbf{Extraction} (645 questions) asks models to locate and directly read out specific information from retrieved pages, while \textbf{Logic} (497 questions) requires further inference or computation over the retrieved content to derive the answer.

\paragraph{MMLongBench.}
MMLongBench~\cite{ma2024mmlongbench} is a multi-modal document understanding benchmark emphasizing long-context perception across heterogeneous content types.
We retain only questions with verifiable reference answers, yielding 847 questions for evaluation.
Questions are tagged by content type—\textbf{Text} (291), \textbf{Table} (217), \textbf{Chart} (178), \textbf{Figure} (290), and \textbf{Layout} (118)—and a single question may carry multiple tags, so the subsets overlap and their counts sum to more than 847.

\section{Implementation Details of the Baseline}\label{app:baselines}

Among all baselines, results for Vanilla RAG, ReAct, Search-R1-VL, and VRAG-RL are taken directly from the VRAG-RL paper~\cite{wang2025vrag}, as our evaluation strictly follows the same experimental protocol.
Results for M3RAG are taken from its original paper~\cite{du2026m3rag}, where the evaluation setup differs from ours (simpler retrieval configuration).
The remaining four baselines—ViDoRAG, EVisRAG, MMSearch-R1, and R1-Router—are reproduced by us under our unified evaluation framework.

\paragraph{Vanilla RAG}
Vanilla RAG is a straightforward retrieval augmented baseline that uses the original question to retrieve relevant pages from the corpus and directly provides the retrieved content to the model for answer generation, without any iterative reasoning or multi-turn interaction.
We follow the visual-based variant, where page images are retrieved via ColQwen2.5-v0.1~\cite{faysse2024colpali} and fed directly to the VLM.

\paragraph{ReAct}
ReAct~\cite{yao2022react} structures the agent's behavior as an interleaved Thought--Action--Observation loop, enabling multi-turn retrieval-augmented reasoning.
At each turn, the model issues a search query based on its current reasoning state and receives a retrieved page image as the observation, continuing until it produces a final answer.

\paragraph{Search-R1-VL}
Search-R1-VL is a visual extension of Search-R1~\cite{jin2025search}, which introduces multi-turn RL-based reasoning into the RAG loop.
The visual variant adapts this framework to image-based retrieval, training on the same dataset and with the same reward and post-processing methods as VRAG-RL, initialized from a cold-start checkpoint.

\paragraph{VRAG-RL}
VRAG-RL~\cite{wang2025vrag} trains an agentic VLM via GRPO-based RL to iteratively retrieve and reason over document page images.
It introduces a crop-and-zoom tool for fine-grained perception and uses trajectory-level rewards that jointly optimize retrieval and answer quality.

\paragraph{M3RAG}
M3RAG~\cite{du2026m3rag} is a multi-agent framework that decomposes the retrieval-reasoning pipeline into specialized agents for multi-modal document understanding.
As its evaluation settings differ from ours in retrieval configuration, we report results directly from the original paper.

\paragraph{ViDoRAG}
ViDoRAG~\cite{wang2025vidorag} adopts an actor-critic multi-agent architecture with separate agents for planning, retrieval, and answer generation, enabling iterative reasoning over visually rich documents. We use ColQwen2.5-v0.1~\cite{faysse2024colpali} as the single-modal search engine, retrieving the top-10 pages per query. The backbone VLM is Qwen2.5-VL-7B. All other components and procedures follow the original ViDoRAG pipeline without modification.

\paragraph{EVisRAG}
EVisRAG~\cite{sun2025visrag} introduces evidence-based reasoning over multiple retrieved images, explicitly extracting and structuring evidence from each retrieved page to support multi-image understanding. We use ColQwen2.5-v0.1~\cite{faysse2024colpali} as the retriever in place of the original VisRAG-Ret, retrieving the top-3 pages per query. The model weights are the officially released EVisRAG-7B. All other settings follow the original EVisRAG configuration.

\paragraph{MMSearch-R1}
MMSearch-R1~\cite{wu2025mmsearch} integrates multimodal search into the reasoning loop, employing cross-modal retrieval mechanisms to fetch contextually aligned information in both visual and textual forms. The original MMSearch-R1 provides two tools: text search and image search (image-to-image retrieval). In our setting, we adapt text search to retrieve images of document pages via ColQwen2.5-v0.1~\cite{faysse2024colpali}. Since image search is not applicable in the document retrieval scenario, invoking it returns a prompt instructing the model to use text search instead. The model weights are the officially released MMSearch-R1-7B. All other settings follow the original configuration.

\paragraph{R1-Router}
R1-Router~\cite{peng2025learning} employs a dynamic routing mechanism trained via Step-GRPO, generating intermediate sub-queries during the reasoning process and directing them selectively to the most appropriate retrieval tool, thereby minimizing unnecessary retrievals while adaptively integrating external evidence. In our setting, all retrieval tools are adapted to retrieve images of document pages via ColQwen2.5-v0.1~\cite{faysse2024colpali}, returning the top-5 pages per query. The maximum number of interaction turns is set to 3. All other settings follow the official R1-Router configuration.

\section{Bbox Case}\label{app:bbox_study}

\begin{figure*}[htbp]
\centering
\includegraphics[width=0.95\textwidth]{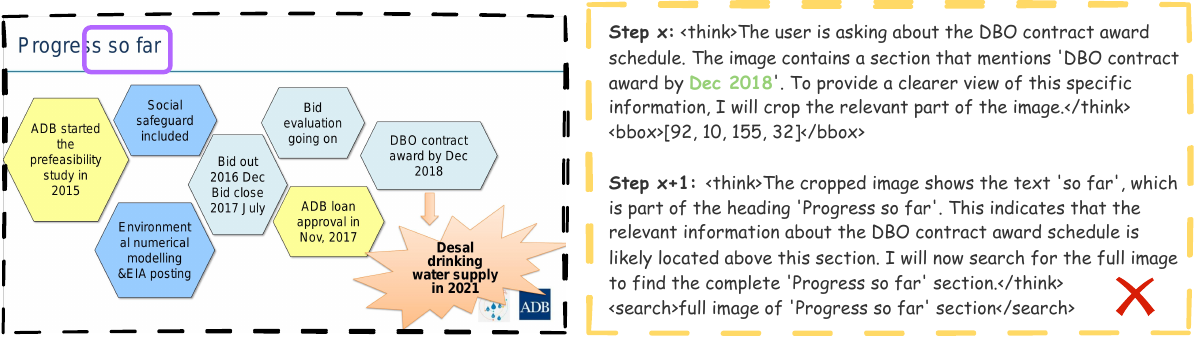}
\caption{A representative case illustrating redundant crop usage in VRAG-RL.}
\label{fig:bbox_case}
\end{figure*}

Figure~\ref{fig:bbox_case} shows a representative failure case of indiscriminate cropping from VRAG-RL.
The question asks about the DBO contract award schedule.
At Step~$x$, the model correctly identifies a relevant section in the retrieved page and issues a \texttt{<bbox>} action targeting the top-left corner of the image.
However, the specified coordinates capture only the heading text ``so far'' (part of ``Progress so far''), missing the actual content entirely.
The resulting cropped patch is uninformative, yet consumes a full interaction turn.
At Step~$x{+}1$, the model recognizes the crop as unhelpful and falls back to issuing a new search query—restarting the retrieval process unnecessarily.

This case illustrates two compounding failure modes: (1) imprecise coordinate prediction that targets an already-legible and non-critical region, and (2) the absence of a correction mechanism, which allows the wasted turn to propagate without recovery guidance.
VISOR addresses both issues via \emph{Visual Action Evaluation}, which gates the crop action before execution, and \emph{Visual Action Correction}, which redirects the model back to its pre-crop reasoning context when a crop proves uninformative.

\section{Case Study}\label{app:case_study}

\begin{figure*}[htbp]
\centering
\includegraphics[width=0.95\textwidth]{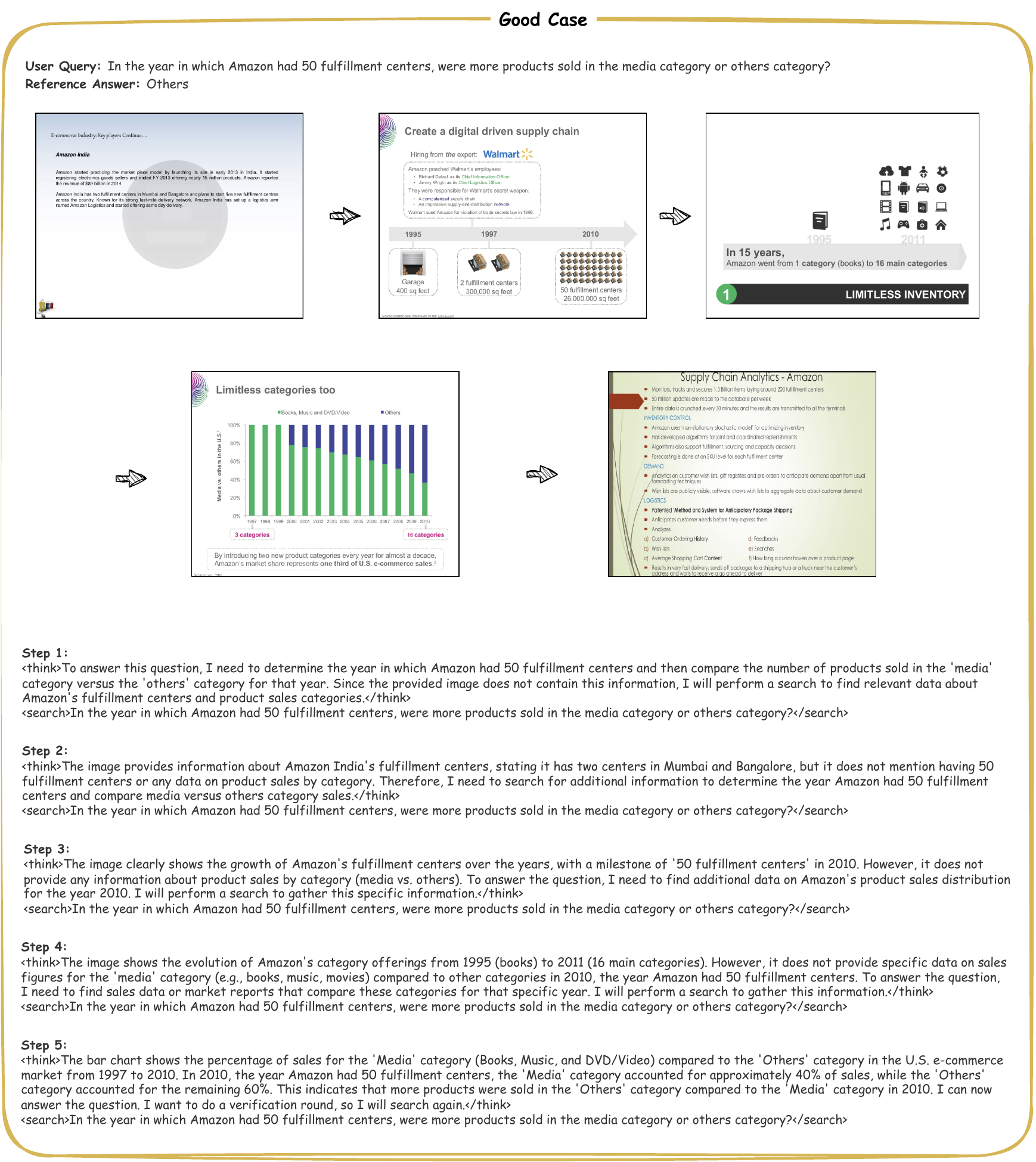}
\caption{A success case of VISOR on a multi-hop SlideVQA question. Retrieved page images are shown at the top for layout reasons; the corresponding reasoning steps are presented below. Step~1 retrieves an irrelevant page from a different document. Step~2 identifies the year 2010 from an Amazon fulfillment center growth chart. Step~3 retrieves a category evolution slide providing context. Step~4 locates the Media vs.\ Others sales breakdown chart, from which the answer is read. The final verification search (Step~5) returns a page with an unrelated chart from a different document, yet VISOR correctly produces the answer from accumulated evidence.}
\label{fig:case_success}
\end{figure*}

Figure~\ref{fig:case_success} shows a representative success case of VISOR on a multi-hop question from SlideVQA:
\textit{``In the year in which Amazon had 50 fulfillment centers, were more products sold in the media category or others category?''}
Answering this question requires a two-step reasoning chain: first locating the year when Amazon reached 50 fulfillment centers, then retrieving sales breakdown data for that specific year.

The trajectory unfolds over five retrieval steps.
At Step~1, the model issues a search and receives an irrelevant page about Amazon India—a different document with no bearing on the question.
The model correctly identifies this as uninformative and continues searching.
At Step~2, VISOR retrieves a timeline slide from the Amazon whitepaper showing that 50 fulfillment centers were reached in \textbf{2010}.
This intermediate finding is distilled into the Evidence Space, bridging the two sub-questions.
At Step~3, VISOR retrieves a category evolution slide covering Amazon's product mix from 1995 to 2011, providing useful context but no sales figures for 2010.
At Step~4, VISOR retrieves a bar chart showing the Media vs.\ Others sales split from 1997 to 2010.
Grounded by the year stored in the Evidence Space, the model directly reads off the 2010 value: Others ($\sim$60\%) $>$ Media ($\sim$40\%), and arrives at the answer.
At Step~5, VISOR performs a verification search and retrieves a page from a different analytics document containing a bar chart—but the chart's content is unrelated to the question.
The model correctly recognizes this page as irrelevant and commits to the answer already derived from accumulated evidence.

Notably, the retrieved pages span \textbf{three distinct source documents}: Steps~2--4 all come from the Amazon whitepaper, while Step~1 and Step~5 each originate from entirely unrelated documents.
This cross-document retrieval is handled transparently by VISOR's per-trajectory deduplication and Evidence Space, which keeps the agent anchored to the original question across all turns regardless of what document each page originates from.
A method without persistent evidence management would risk losing the intermediate finding (2010) as irrelevant pages accumulate, leading to search drift or an incorrect answer.

\begin{figure*}[htbp]
\centering
\includegraphics[width=\textwidth]{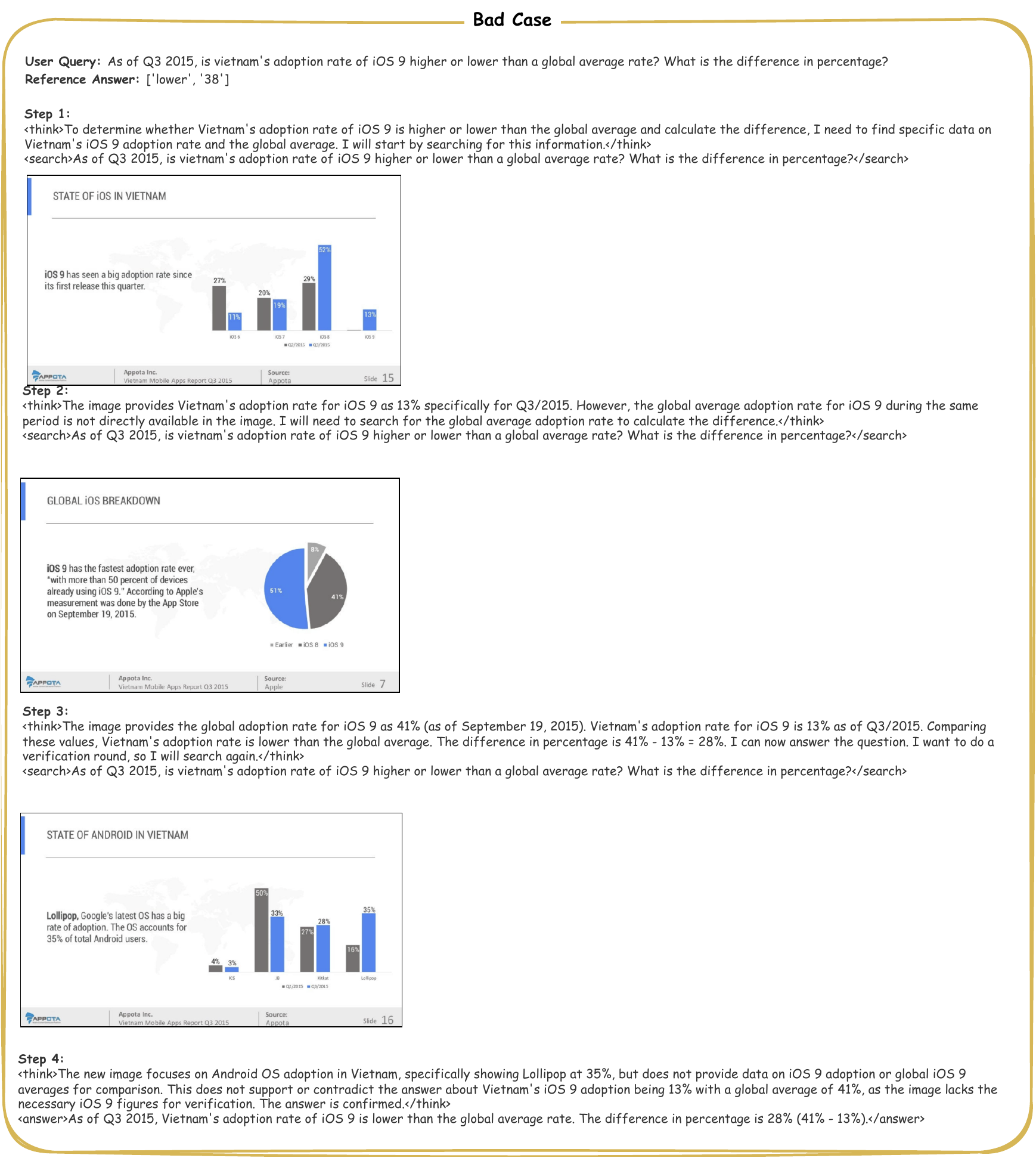}
\caption{A failure case of VISOR on a multi-hop MMLongBench question. Both reference pages are retrieved correctly, but the model misreads the legend of a pie chart, confusing the iOS~8 slice (gray, 41\%) for iOS~9 (blue, 51\%), leading to an incorrect final answer despite successful retrieval.}
\label{fig:case_failure}
\end{figure*}

Figure~\ref{fig:case_failure} illustrates a representative failure case of VISOR on a multi-hop question from MMLongBench:
\textit{``As of Q3 2015, is Vietnam's adoption rate of iOS~9 higher or lower than the global average rate? What is the difference in percentage?''}
This question requires retrieving two separate slides from the same report and combining a country-level bar chart with a global pie chart.

VISOR correctly retrieves both reference pages.
At Step~1, the model fetches the Vietnam iOS breakdown slide, which contains a grouped bar chart clearly annotated with per-version adoption rates.
The model accurately reads Vietnam's iOS~9 adoption for Q3/2015 as \textbf{13\%}.
At Step~2, the model retrieves the global iOS breakdown slide, which shows a three-sector pie chart with iOS~9 (blue, \textbf{51\%}), iOS~8 (gray, \textbf{41\%}), and Earlier (light gray, 8\%).
However, the model misidentifies the sectors: it reads the \emph{gray} iOS~8 sector (41\%) as the iOS~9 share, overlooking the legend.
As a result, the model computes $41\% - 13\% = 28\%$ instead of the correct $51\% - 13\% = 38\%$.
The directional answer (``lower'') is correct, but the numerical answer is wrong due to this legend misinterpretation.

\begin{figure*}[htbp]
\centering
\includegraphics[width=\textwidth]{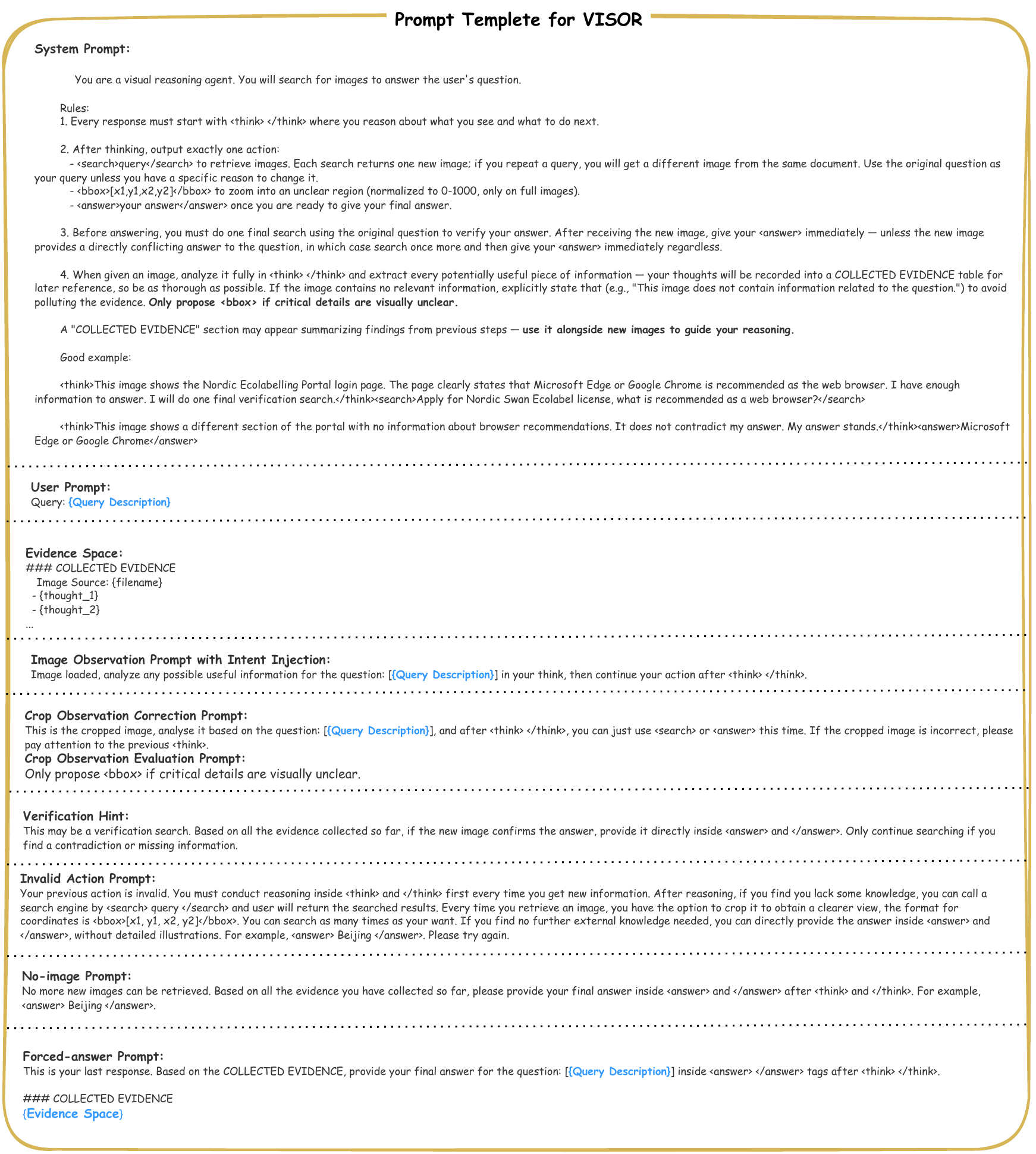}
\caption{All prompt templates used in VISOR. Each template corresponds to a distinct interaction event in the agent loop.}
\label{fig:prompts}
\end{figure*}

This failure mode cannot be remedied by better retrieval or longer reasoning: the bottleneck is purely visual—the model fails to correctly associate legend colors with the corresponding pie sectors, a task that requires precise fine-grained perception of small-font legend entries alongside visually similar chart elements.
It highlights an inherent limitation of the current 7B backbone and suggests that improved visual grounding capabilities in future VLMs would directly translate to higher benchmark performance.

\section{Prompt Templates}\label{app:prompts}

All prompt templates used in VISOR are illustrated in Figure~\ref{fig:prompts}.
The prompts described below are designed for the 7B backbone (Qwen2.5-VL-7B); for smaller models such as the 3B variant, we lightly simplify the wording to reduce comprehension burden and improve format adherence.

\textbf{System Prompt.}
Injected once at the start of each trajectory, the System Prompt defines the agent's action space and behavioral rules.
It specifies three valid actions: \texttt{<search>} to retrieve a new page, \texttt{<bbox>} to zoom into an unclear region, and \texttt{<answer>} to produce a final response.
Crucially, it mandates that every response begin with a \texttt{<think>} block, and requires the agent to perform one final verification search before committing to an answer.
It also instructs the agent to extract all potentially useful information from each retrieved image into its \texttt{<think>}, as these thoughts are recorded into the persistent Evidence Space for later reference.

\textbf{User Prompt.}
Appended immediately after the System Prompt, the User Prompt provides the original question for the current trajectory.

\textbf{Evidence Space.}
The Evidence Space is a structured ledger maintained across all turns.
After each retrieval step, the agent's \texttt{<think>} content is extracted and appended to the Evidence Space under the corresponding image source filename.
The accumulated Evidence Space is then injected into subsequent prompts (e.g., the Image Observation Prompt and Forced-answer Prompt) so that the agent can reason over all prior findings without relying on raw conversation history.

\textbf{Image Observation Prompt.}
Appended after each retrieved page image, this prompt embeds the \emph{Intent Injection}—repeating the original question—to keep the agent anchored to the user's query regardless of the retrieved content.
It instructs the agent to analyze the image and record findings in \texttt{<think>} before issuing the next action.

\textbf{Crop Observation Evaluation Prompt.}
This prompt is appended alongside the Image Observation Prompt and serves as the \emph{Visual Action Evaluation} gate: it reminds the agent to propose \texttt{<bbox>} only if critical details are visually unclear, discouraging unnecessary crop actions on already legible regions.

\textbf{Crop Observation Correction Prompt.}
Returned after a \texttt{<bbox>} action along with the cropped image, this prompt implements \emph{Visual Action Correction}: it constrains the agent's next action to either \texttt{<search>} or \texttt{<answer>}, preventing chained crops, and instructs the agent to refer back to its previous \texttt{<think>} if the cropped region turns out to be uninformative.

\textbf{Verification Hint.}
Injected when the system detects that the current retrieval step may be a verification search (triggered by the agent's prior \texttt{<think>} signaling readiness to confirm), this prompt instructs the agent to issue a final \texttt{<answer>} immediately if the new image confirms the accumulated evidence, and to continue searching only if a direct contradiction is found.

\textbf{No-image Prompt.}
Triggered when all top-$k$ retrieved pages have already been shown to the agent in prior turns, this prompt notifies the agent that no new pages are available and directs it to produce a final answer solely from the accumulated Evidence Space.

\textbf{Forced-answer Prompt.}
Injected at the final turn when the maximum number of turns is reached, this prompt requires the agent to produce an answer unconditionally, providing the full Evidence Space as context to ground the response.

\textbf{Invalid Action Prompt.}
Returned whenever the agent produces a malformed output (\emph{e.g.}, missing \texttt{<think>} tags, an unrecognized action token, or out-of-range \texttt{<bbox>} coordinates), this prompt describes the correct format and instructs the agent to retry.

\section{Sensitivity Analysis of Sliding Window Size}\label{app:window_sensitivity}

In the Dynamic Trajectory mechanism, the sliding window size $W$ controls how many recent interaction turns are retained in the live context at each step.
A smaller $W$ bounds the number of in-context image tokens but risks discarding recent raw observations, while a larger $W$ preserves more interaction history at the cost of increased token consumption.
We empirically set $W = 2$ in all main experiments; here we investigate how sensitive VISOR's performance is to this choice.

\paragraph{Experimental Setup.}
We vary $W \in \{1, 2, 3, 4\}$ and evaluate the fine-tuned VISOR (Qwen2.5-VL-7B, SFT+RL) on SlideVQA (2,215 test questions).
All other settings are identical to the main experiments.
Results are reported in Table~\ref{tab:window_sensitivity}.

\begin{table}[h]
\centering
\caption{Sensitivity of VISOR to sliding window size $W$ on the fine-tuned model (SlideVQA). We report overall accuracy (\%).}
\label{tab:window_sensitivity}
\begin{tabular}{cc}
\hline
\textbf{Window Size $W$} & \textbf{SlideVQA} \\
\hline
1 & 70.43 \\
2 & \textbf{72.37} \\
3 & 71.56 \\
4 & 68.17 \\
\hline
\end{tabular}
\end{table}

\paragraph{Analysis.}
VISOR is largely insensitive to $W$ for small values ($W \leq 3$), with a performance gap of less than 2 percentage points relative to the optimal setting.
This robustness stems from the complementary role of the Evidence Space (ES): even when earlier turns are evicted from the raw context window, their distilled evidence is preserved in $\mathcal{E}_t$ and re-injected at every subsequent turn, so the agent does not suffer a hard information loss.

$W = 1$ causes a mild accuracy drop (70.43\% vs.\ 72.37\%).
When the agent issues a \texttt{<bbox>} crop action, a window of size~1 retains \emph{only} the crop observation, discarding the original full-page image that motivated the crop.
The agent can no longer directly compare the cropped detail with its broader page context, slightly impairing fine-grained reasoning.
$W = 2$ recovers this capability: the preceding retrieve turn (full-page image plus initial reasoning) is kept alongside the crop result, giving the agent a complete \textit{retrieve-then-crop} reasoning chain.

Increasing $W$ to 4 leads to a more substantial drop (68.17\%), approaching the ablation result of removing SW entirely (67.54\%).
With a larger window, older page images that are no longer actionable for the current turn are kept in the live context, introducing additional visual noise and increasing token consumption without proportional benefit.
Taken together, $W = 2$ strikes the best balance between context efficiency and information completeness, and we adopt this setting throughout the paper.

\end{document}